\documentclass[runningheads]{llncs}

\usepackage{eccv}

\usepackage{eccvabbrv}

\usepackage{graphicx}
\usepackage{booktabs}

\usepackage[accsupp]{axessibility}  

\usepackage{algorithm}
\usepackage[noend]{algpseudocode} 

\usepackage{hyperref}

\usepackage{orcidlink}

\begin{document}

\title{DPAC: Distribution-Preserving Adversarial Control for Diffusion Sampling} 


\author{
    Han--Jin Lee\inst{1}\orcidlink{0009--0009--0826--9833} \and
    Han--Ju Lee\inst{1}\orcidlink{0009--0009--6316--7354} \and
    Jin--Seong Kim\inst{1}\orcidlink{0009--0008--8748--9676} \and
    Seok--Hwan Choi\inst{1}
}

\authorrunning{F.~Author et al.}

\institute{Yonsei University, Wonju, South Korea\\
\email{\{han--.--jin,hanleju,js\_kim,sh.choi\}@yonsei.ac.kr}}

\maketitle
\begin{abstract}
    Adversarially guided diffusion sampling often achieves the target class, but sample quality degrades as deviations between the adversarially controlled and nominal trajectories accumulate.
    We formalize this degradation as a \emph{path-space Kullback-Leibler divergence}(path-KL) between controlled and nominal (uncontrolled) diffusion processes, thereby showing via Girsanov’s theorem that it exactly equals the control energy.
    Building on this stochastic optimal control (SOC) view, we theoretically establish that minimizing this path-KL simultaneously tightens upper bounds on both the 2-Wasserstein distance and Fréchet Inception Distance (FID), revealing a principled connection between adversarial control energy and perceptual fidelity.
    From a variational perspective, we derive a first-order optimality condition for the control: among all directions that yield the same classification gain, the component tangent to iso-(log-)density surfaces (i.e., orthogonal to the score) minimizes path-KL, whereas the normal component directly increases distributional drift.
    This leads to \textbf{DPAC} (Distribution-Preserving Adversarial Control), a diffusion guidance rule that projects adversarial gradients onto the tangent space defined by the generative score geometry.
    We further show that in discrete solvers, the tangent projection cancels the $O(\Delta t)$ leading error term in the Wasserstein distance, achieving an $O(\Delta t^2)$ quality gap; moreover, it remains second-order robust to score or metric approximation.
    Empirical studies on ImageNet-100 validate the theoretical predictions, confirming that DPAC achieves lower FID and estimated path-KL at matched attack success rates. 
\end{abstract}
\section{Introduction}
\label{sec:intro}

Denoising diffusion models~\cite{ho_ddpm_2020, song_score_2021} represent the state-of-the-art in generative modeling~\cite{robin_high_2022, dhariwal_diffusion_2021}, largely due to their amenability to guidance~\cite{ho_classifier_2022}.
This control mechanism can steer generation towards prompts or class labels, and can also be used for evaluating model robustness by generating unrestricted adversarial examples (UAEs)~\cite{dai_advdiff_2024, chen_advdiffuser_2023, xue_diffusion-based_2023}.

However, existing gradient-based guidance methods like AdvDiff~\cite{dai_advdiff_2024} suffer from a fundamental, unresolved flaw.
As guidance strength increases to maximize the Attack Success Rate (ASR), the sample quality catastrophically collapses.
The resulting high-FID, artifact-laden images are invalid as ``adversarial examples,'' which must be both effective (high ASR) and realistic (low FID).

In this paper, we diagnose and explain and mitigate this instability.
We first identify that the quality collapse is caused by a ``normal'' (score-parallel) component of the guidance gradient.
This component, while effective for the adversarial task, aggressively pushes the sampling trajectory off the data manifold.
Based on this diagnosis, we propose \textbf{DPAC (Distribution-Preserving Adversarial Control)}, a new guidance framework built on the principle of \emph{tangential control}.
DPAC uses a geometric projection $\Pi_\perp$ to surgically remove this harmful normal component, isolating the ``tangential'' gradient that steers generation within the data manifold.

The results show that DPAC substantially mitigates the core ASR–FID trade-off: at high guidance strengths where AdvDiff's quality catastrophically collapses (FID 69.37), DPAC remains stable and avoids catastrophic artifacts/collapse (FID 44.89).
%
%
Moreover, our method is significantly more efficient.
DPAC achieves a superior peak fidelity (FID 33.90) using only one-third the energy (guidance strength) of the baseline's inferior optimum (FID 34.66).

\section{Background}
\label{sec:background}

\subsection{Diffusion Models as SDEs}
Denoising diffusion models~\cite{song_score_2021, song_ddim_2021, ho_ddpm_2020} define a forward noising process that gradually perturbs data $x_0 \sim p_{\text{data}}$ into noise via a Stochastic Differential Equation (SDE):
\begin{equation}
    dX_t = f(X_t, t)\,dt + g(t)\,dW_t, \quad t \in [0, 1].
\label{eq:fwd_sde}
\end{equation}
Here, $f(\cdot, t)$ is a linear drift, $g(t)$ is the diffusion coefficient, and $W_t$ is a standard Wiener process.
As $t \to 1$, $X_t$ converges to a simple prior, typically $\mathcal{N}(0, I)$.

A key insight~\cite{aenderson_reverse_1982, song_score_2021} is that this process is reversible.
We can generate new data by solving the corresponding reverse-time SDE from $t=1$ to $t=0$:
\begin{equation}
    dX_t = [f(X_t, t) - g(t)^2 s_\theta(X_t, t)]\,dt + g(t)\,d\bar{W}_t,
\label{eq:rev_sde}
\end{equation}
where $d\bar{W}_t$ is a reverse-time Wiener process.
The only unknown term is the \textbf{score function} $s_\theta(X_t, t) = \nabla_{X_t} \log p_t(X_t)$, which points in the direction of highest data density.
This function is approximated by a time-conditioned neural network $s_\theta$ (e.g., a U-Net) trained via denoising score-matching~\cite{ho_ddpm_2020}.

\subsection{Adversarial Control and its Instability}
The reverse SDE provides a natural mechanism for control.
An adversarial attack (or UAE generation) steers the sampling process to fool a target classifier $v_\phi$.
This is achieved by defining a controlled SDE (as detailed in \cref{sec:theory}, \cref{eq:sde_control}), which injects an additional control drift $u_t$:
\begin{equation}
    dX_t = [\text{Original Drift}] + g(t) u_t(X_t, t)\,dt + g(t)\,d\bar{W}_t.
\label{eq:controlled_sde_bg}
\end{equation}
This SDE formulation provides a theoretical lens to interpret existing heuristic methods. For example, the guidance strategy used in AdvDiff~\cite{dai_advdiff_2024} is equivalent to setting the control $u_t$ as the raw classifier gradient, $u_t \propto \nabla_{X_t}  \ell_{\text{tar}}(v_\phi(X_t); y_{\text{tar}})$, where $y_{\text{tar}}$ is the target class and $ \ell_{\text{tar}}$ is the attack objective function (e.g., cross-entropy loss).
This successfully increases the ASR.

However, this raw gradient control is the source of the instability.
As we formalize in \cref{sec:theory}, any gradient vector $u_t$ can be decomposed into two components relative to the data manifold at $X_t$:
(1) a \textbf{tangential component} ($u_\perp$), which steers the sample along the iso-density surface (preserving the density $p_t$), and (2) a \textbf{normal component} ($u_\parallel$), which is parallel to the score $s_\theta$ and steers the sample \emph{off} the manifold (distorting the density $p_t$).
Standard guidance (e.g., AdvDiff) uses this raw vector $u_t$ indiscriminately, thus injecting the harmful normal component $u_\parallel$.
This distortion explicitly pushes the sampling trajectory away from the true data distribution, which manifests as the catastrophic FID collapse observed in our experiments (\cref{fig:main_quantitative_results}a).
Our work (\cref{sec:method}) is motivated by the need to surgically remove this harmful component.
\section{Theoretical Foundations}
\label{sec:theory}

\subsection{Reverse SDE Formulation}
We begin by formalizing the concepts from~\cref{sec:background}.
Starting from the score-based reverse SDE (\cref{eq:rev_sde}) and the general control framework (\cref{eq:controlled_sde_bg}),
we define our adversarially guided sampling process as:
\begin{equation}
dX_t = [f(X_t,t) - g(t)^2\,s_\theta(X_t,t\mid y_{\mathrm{gt}})]\,dt
+ g(t)\,u_t(X_t,t)\,dt + g(t)\,d\bar{W}_t,
\label{eq:sde_control}
\end{equation}
where $s_\theta(\cdot,t\mid y_{\mathrm{gt}})$ is the clean-conditioned score function for a ground-truth label $y_{\mathrm{gt}}$~\cite{ho_classifier_2022},
often approximated in practice with classifier-free guidance (CFG), and $u_t$ is the adversarial control.

The path distributions of the uncontrolled ($u_t\equiv0$) and controlled processes are denoted by $\mathcal{P}^0$ and $\mathcal{P}^u$, respectively.
The corresponding marginal densities at each time $t$ are written $p_t^0$ and $p_t^u$.
The control term $g(t)u_t\,dt$ perturbs the reverse diffusion through its drift, and this specific scaling by $g(t)$
yields a convenient expression for the induced path-space divergence via Girsanov's Theorem~\cite{kallianpur_girsanov_2000},
which we use throughout to quantify how strongly guidance distorts the sampling distribution.
Intuitively, $u_t$ injects a continuous steering signal that shifts the sampling trajectory toward a target label $y_{\mathrm{tar}}$,
while still evolving under the same diffusion noise and the clean-conditioned score field.

We assume standard regularity (e.g., $g(t)>0$, Lipschitz/linear-growth of $f$ and $s_\theta$, and square-integrable progressively measurable control $u_t$) so the reverse SDE is well-posed and Girsanov applies (Novikov).
For brevity, write $s_t:=s_\theta(x,t\mid y_{\mathrm{gt}})$ and $g_t:=g(t)$.

\subsection{Path-Space Divergence and Perceptual Bounds}
\label{sec:path-space-divergence}
To quantify how strongly the control $u_t$ perturbs the reverse-time diffusion process, we analyze the KL divergence between the path distributions $\mathcal{P}^0$ and $\mathcal{P}^u$.
By Girsanov’s theorem~\cite{kallianpur_girsanov_2000}, under Novikov’s condition, the relative entropy between them equals the cumulative control energy:
\begin{equation}
KL(\mathcal{P}^u\Vert \mathcal{P}^0)=\tfrac12\,\mathbb{E}_{\mathcal{P}^u}\!\int_0^1\|u_t(X_t,t)\|_2^2\,dt.
\label{eq:girsanov}
\end{equation}
Hence, the path-KL directly measures the total energy injected by the control.
Because it aggregates the squared deviation introduced in the drift over the entire sampling horizon, it provides a principled measure of distributional distortion induced by guidance.
(The equivalence in \cref{eq:girsanov} relies on the drift scaling by $g(t)$ in \cref{eq:sde_control} and the nondegeneracy condition $g(t)\!>\!0$.)

The path-space divergence also upper-bounds the discrepancy of terminal distributions.
The marginalization inequality
\begin{equation}
KL(p_t^u\Vert p_t^0)\le KL(\mathcal{P}^u\Vert \mathcal{P}^0),\quad \forall t\in[0,1],
\label{eq:data_proc}
\end{equation}
implies that controlling the path-KL implicitly limits the KL divergence at every time step.
If the clean terminal density $p_0^0$ satisfies the Talagrand $T_2(C)$ inequality~\cite{molly_talagrand_2002}, then the Wasserstein distance obeys
\begin{equation}
W_2^2(p_0^u,p_0^0)\le 2C\,KL(p_0^u\Vert p_0^0)
\le 2C\,KL(\mathcal{P}^u\Vert \mathcal{P}^0).
\label{eq:talagrand}
\end{equation}
Moreover, let $\phi$ be the $L$-Lipschitz feature embedding used for FID calculation~\cite{heusel_GANs_2017}
(hence $W_2(\phi_\#p,\phi_\#q)\le L\,W_2(p,q)$).
Letting $\varepsilon_{\text{gauss}}$ denote the Gaussian approximation error, the triangle inequality combined with the $L$-Lipschitz property of $\phi$
yields an upper bound on the square root of the FID:
\begin{equation}
\begin{aligned}
\sqrt{FID(\phi_\#p_0^u,\phi_\#p_0^0)}
&\le L\,W_2(p_0^u,p_0^0) + \varepsilon_{\text{gauss}} \\
&\le \underbrace{L\sqrt{2C}}_{=:\tilde K}\,\sqrt{KL(\mathcal{P}^u\Vert \mathcal{P}^0)} + \varepsilon_{\text{gauss}}.
\end{aligned}
\label{eq:fid}
\end{equation}
Thus, \cref{eq:data_proc} and \cref{eq:fid} link control energy ($KL(\mathcal{P}^u\Vert \mathcal{P}^0)$) with perceptual fidelity (FID).
In the remainder of the paper, we treat the path-KL as a distortion metric and design guidance that achieves the adversarial objective without injecting unnecessary energy.
This motivates the decomposition and projection strategy introduced in \cref{sec:method}.

\subsection{Tangential Control and First-Order Optimality}
\label{sec:tangential}

We now investigate how to design an adversarial control $u_t$ that minimally changes the data distribution while maximizing the target classification gain.
The Fokker--Planck equation associated with \cref{eq:sde_control} is~\cite{Oksendal_SDE_1992, song_score_2021}
\begin{equation}
\partial_t p_t=-\nabla\!\cdot(\mu p_t)-\nabla\!\cdot(g_t u_t p_t)
+\tfrac12\nabla\!\cdot(g_t^2\nabla p_t),
\label{eq:fp}
\end{equation}
where $\mu=f-g^2 s$.
It is convenient to separate the uncontrolled operator $\mathcal{L}_t^\ast p := -\nabla\!\cdot(\mu p) + \tfrac12 g_t^2 \Delta p$ from the control-induced flux:
\begin{equation}
\partial_t p_t = \mathcal{L}_t^\ast p_t - g_t\,\nabla\!\cdot(p_t u_t).
\label{eq:fp_split}
\end{equation}

Let $p_t$ denote the current marginal density used to define the local density manifold (in theory, the exact $p_t$; in practice, approximated through the score $s_t$).
Consider a small perturbation of the control $u_t \mapsto u_t + \varepsilon v_t$ and define the G\^ateaux derivative
$\delta p_t := \left.\tfrac{d}{d\varepsilon}p_t^{(u+\varepsilon v)}\right|_{\varepsilon=0}$.
The control enters the Fokker--Planck equation only through the divergence term, and the linearization yields the forcing
\begin{equation}
\partial_t(\delta p_t)=\mathcal{L}_t^\ast(\delta p_t) - g_t\,\nabla\!\cdot(p_t v_t)
\quad(\text{up to terms linear in } \delta p_t).
\label{eq:fp_lin}
\end{equation}
Thus, directions satisfying $\nabla\!\cdot(p_t v_t)=0$ produce no \emph{first-order} density change through the control channel (the control-induced flux is divergence-free w.r.t.\ $p_t$).
We refer to this subspace as the tangential (distribution-preserving) space:
\begin{equation}
\mathcal{T}_t := \{v:\nabla\!\cdot(p_t v)=0\}.
\label{eq:tangent_space}
\end{equation}

Formally, based on a weighted Hodge--Helmholtz decomposition~\cite{bhatia_helmholtz_2013}, any vector field $u_t$ can be uniquely decomposed (w.r.t.\ $p_t$) as
\begin{equation}
u_t=\nabla\varphi_t+v_t,\qquad v_t\in\mathcal{T}_t,
\label{eq:hodge}
\end{equation}
where $\nabla\varphi_t$ is the normal (density-changing) component and $v_t$ is the tangential (density-preserving) component.
These components are orthogonal under the $L^2(p_t,G_t)$ inner product
$\langle v,w\rangle_{L^2(p_t,G_t)}=\mathbb{E}_{p_t}[v^\top G_t w]$.
For full consistency with the metric $G_t$, the divergence-free subspace can be defined by $\{v:\nabla\!\cdot(p_t G_t v)=0\}$; we use \cref{eq:tangent_space} for clarity, and the two coincide when $G_t\approx I$.

Let $\ell_{\mathrm{tar}}(x_0;y_{\mathrm{tar}})$ denote the target classification loss at the terminal time $t=0$.
To increase the expected objective $\mathbb{E}[\ell_{\mathrm{tar}}(X_0)]$, we compute its first-order variation under a small control perturbation using an adjoint argument~\cite{chen_neural_2018}.
Define the backward potential $\psi(x,t)$ as the solution to the backward Kolmogorov equation under the \emph{uncontrolled} reverse drift $\mu$:

\begin{equation}
-\partial_t \psi=\mu\!\cdot\!\nabla\psi+\tfrac12 g_t^2\,\Delta\psi,\quad \psi(\cdot,0)=\ell_{\mathrm{tar}}(\cdot).
\label{eq:kolmogorov}
\end{equation}

Equivalently, $\psi(x,t)=\mathbb{E}[\ell_{\mathrm{tar}}(X_0)\mid X_t=x]$ for the nominal process, hence $\nabla\psi$ encodes the local sensitivity of the terminal loss to state perturbations.
This yields the sensitivity field $h_t := g_t\nabla\psi(x,t)$ and the first-order variation

\begin{equation}
\delta\mathbb{E}[\ell_{\mathrm{tar}}(X_0)]
=\varepsilon\!\int_0^1\!\mathbb{E}\big[\langle h_t,v_t\rangle\big]dt.
\label{eq:adjoint}
\end{equation}

We now enforce distribution preservation by restricting to $u_t\in \mathcal{T}_t$, and seek the least-energy control that achieves a prescribed first-order gain.
Specifically, we minimize the total energy $\frac12\int_0^1\|u_t\|_{L^2(p_t,G_t)}^2dt$
subject to a fixed gain $\int_0^1\langle h_t,u_t\rangle_{L^2(p_t,G_t)}dt=\Gamma$
and the constraint $u_t\in\mathcal{T}_t$.
The unique minimizer is the orthogonal projection (in $L^2(p_t,G_t)$) of $G_t^{-1}h_t$ onto $\mathcal{T}_t$:
\begin{equation}
u_t^{\mathrm{tan}}=\Pi_{\mathcal{T}_t}^{G_t}(G_t^{-1}h_t).
\label{eq:proj_tan}
\end{equation}
In this constrained problem, the normal component contributes only additional energy without providing tangential (distribution-preserving) gain, so the projection in \cref{eq:proj_tan} is the first-order optimal solution.

In high dimensions, the exact $p_t$-divergence-free projector in \cref{eq:proj_tan} is computationally intractable.
We thus approximate it by removing only the component along the score direction $s_t$.
Define the parallel ($\Pi_\parallel$) and perpendicular ($\Pi_\perp$) projections relative to a vector $s$ and metric $G$:
\begin{equation}
\Pi_\parallel^{(s,G)}u:=\frac{\langle u,s\rangle_G}{\langle s,s\rangle_G}s,\qquad
\Pi_\perp^{(s,G)}u:=u-\Pi_\parallel^{(s,G)}u,
\label{eq:proj_operators}
\end{equation}
and set $u_t^{\mathrm{proj}} = \Pi_\perp^{(s_t, G_t)} w_t$ with $w_t := G_t^{-1}h_t$:
\begin{equation}
u_t^{\mathrm{proj}}
= w_t - \frac{\langle w_t,s_t\rangle_{G_t}}{\langle s_t,s_t\rangle_{G_t}}\,s_t,
\label{eq:score_proj}
\end{equation}
where $\langle a,b\rangle_{G_t}:=a^\top G_t b$.
Exact density preservation is achieved by \cref{eq:proj_tan}; \cref{eq:score_proj} is a computationally efficient first-order surrogate that removes the score-parallel (density-changing) component emphasized in our analysis.

The path-energy viewpoint in \cref{eq:girsanov} is closely related to SOC formulations that optimize control energy together with a terminal objective.
For example, Path Integral Sampler casts sampling as an SOC problem with running cost given by control energy and derives a $\nabla\log(\cdot)$-type optimal controller via a path-integral representation~\cite{zhang2021path},
and Stochastic Control Guidance applies an SOC-inspired, plug-and-play guidance mechanism to handle non-differentiable rules in diffusion models~\cite{huang_scg_rule_guided_diffusion}.
Our derivation here focuses on a complementary constrained formulation: for a prescribed first-order gain, the tangential projection yields the minimum-energy, distribution-preserving direction, motivating the practical projector used in DPAC.


\subsection{Discrete Bound and Robustness}
\label{sec:discrete_robust}

We now examine the implications of this structure for discrete samplers such as DDIM~\cite{song_ddim_2021} or DDPM~\cite{ho_ddpm_2020}.
We adopt a shorthand for discrete time steps, where the subscript $k$ implies evaluation at $(x_k,t_k)$.
Let $\mu_k := \mu(x_k, t_k)$, $g_k := g(t_k)$, $u_k := u_{t_k}$, with step size $\Delta t_k$; we adopt shared-noise coupling:
\begin{equation}
x_{k-1} = x_k + \mu_k \Delta t_k + g_k u_k \Delta t_k + g_k \sqrt{\Delta t_k}\,\epsilon_k,
\label{eq:discrete}
\end{equation}
where $\epsilon_k \sim \mathcal{N}(0,I)$.
We write $\Delta t_{\max}:=\max_k\Delta t_k$, and assume synchronous coupling (the same $\epsilon_k$) for the controlled and nominal chains.
Let $\hat p_0^u$ and $\hat p_0^0$ denote the terminal distributions of the discretized controlled and nominal chains, respectively.

Under Lipschitz continuity of $f$ and $s$ and uniform conditioning bounds on $G_t$, the terminal discrepancy satisfies
\begin{equation}
\begin{split}
W_2^2(\hat p_0^u,\hat p_0^0)
\le C_1 \sum_{k=1}^K \|\Pi_\parallel^{(s_k, G_k)} u_k\|_{G_k}^2 \Delta t_k
+ C_2 \Delta t_{\max}^2,
\end{split}
\label{eq:w2_bound}
\end{equation}
for constants $C_1,C_2$ depending only on the Lipschitz and conditioning bounds.
This form is aligned with SOC-based sampling analyses that quantify the impact of suboptimal control and time discretization on terminal transport error (e.g., Wasserstein bounds in Path Integral Sampler)~\cite{zhang2021path}.

The key implication of \cref{eq:w2_bound} is structural: the dominant, step-size invariant term is driven entirely by the score-parallel component $\Pi_\parallel^{(s_k,G_k)}u_k$, whereas tangential (score-orthogonal) control cancels it and leaves only the $O(\Delta t_{\max}^2)$ discretization remainder.
Accordingly, distribution-preserving control can improve terminal fidelity by removing the non-vanishing leading term in \cref{eq:w2_bound}, although the bound itself is not intended to be numerically tight.

Using synchronous coupling of the controlled and nominal chains with shared noise, let $\delta_k := x_k^u - x_k^0$.
A one-step expansion gives
$\delta_{k-1}=\delta_k+(\mu_k^u-\mu_k^0)\Delta t_k+g_k u_k\Delta t_k$,
where $\mu_k^u:=\mu(x_k^u,t_k)$ and $\mu_k^0:=\mu(x_k^0,t_k)$.
Under Lipschitz continuity of $\mu$, the drift difference satisfies $\|\mu_k^u-\mu_k^0\|\le L\|\delta_k\|$,
so the first-order forcing in $\delta_{k-1}$ arises from $g_k u_k\Delta t_k$.
Decomposing $u_k$ into score-parallel and score-orthogonal parts, the score-orthogonal component contributes only at higher order under the coupling used in the bound,
leading to \cref{eq:w2_bound} (see supplementary materials for a detailed proof).

Finally, we state a robustness bound for the projected control.
When the score, metric, or sensitivity fields are perturbed by bounded estimation errors $\varepsilon_t$ (score), $\Delta_t$ (metric), and $e_t$ (sensitivity),
the resulting increase in path energy remains second-order:
\begin{equation}
\begin{split}
KL(\mathcal{P}^{\tilde u^{\mathrm{tan}}}\Vert \mathcal{P}^0)
 - KL(\mathcal{P}^{u^{\mathrm{tan}}}\Vert \mathcal{P}^0)
\le K \!\int_0^1\!
(\|\varepsilon_t\|^2 + \|\Delta_t\|^2 + \|e_t\|^2)\,dt,
\end{split}
\label{eq:robust}
\end{equation}
where $K$ depends only on the Lipschitz constants of $(f,s)$ and spectral bounds of $G_t$.
This follows from the strong convexity of the energy functional
$\mathcal{E}(u)=\tfrac{1}{2}\!\int\!\|u_t\|_{L^2(p_t,G_t)}^2\,dt$,
which makes the tangential projection $u_t^{\mathrm{tan}}=\Pi_{\mathcal{T}_t}^{G_t}(G_t^{-1}h_t)$ non-expansive.
Perturbing the score, metric, or sensitivity fields by $(\varepsilon_t,\Delta_t,e_t)$ inflates the energy by at most a constant multiple of their squared norms
(via a Pythagorean/Bregman inequality), yielding \cref{eq:robust} (see supplementary materials for a detailed proof).

Together, these results support the principle that minimizing the path-space KL yields guidance that achieves the target objective while preserving visual fidelity,
and they motivate the practical algorithm described in \cref{sec:method}.
\section{Method: DPAC}
\label{sec:method}

\subsection{The Practical DPAC Projection Rule}
\label{sec:method_practical_rule}
Our method, DPAC (Distribution-Preserving Adversarial Control), implements the principle from \cref{sec:theory} with a computationally efficient surrogate.
In \cref{sec:tangential}, we motivated that, under a fixed first-order adversarial gain, the minimum-energy direction is obtained by restricting the control to distribution-preserving (tangential) directions and projecting out density-changing components.
However, the ideal projector onto the tangential subspace and the exact sensitivity field are intractable in high dimensions.

We therefore make two practical approximations.
First, we approximate the theoretical sensitivity with a one-step lookahead gradient $w_k$, computed via a discrete sensitivity oracle $\mathcal{G}_{k-1}$:
\begin{equation}
 w_k = \mathcal{G}_{k-1}(x_{k-1}^{\text{clean}})
 := \nabla_{x_{k-1}^{\text{clean}}} \ell_{\mathrm{tar}}(x_{k-1}^{\text{clean}}),
\label{eq:pgd_gradient}
\end{equation}
where $\ell_{\mathrm{tar}}$ is the target loss.
This gradient is computed only on steps where the guidance scale $\eta_k$ is active.

Second, we implement a pointwise surrogate of tangential control by removing the score-parallel component from $w_k$ using the metric-weighted inner product:
\begin{equation}
u_k^{\star} = w_k - \frac{\langle w_k, s_k\rangle_{G_k}}{\langle s_k, s_k\rangle_{G_k}+\epsilon}s_k,
\label{eq:score_proj_method}
\end{equation}
with a tiny $\epsilon\!\sim\!10^{-8}$ for numerical stability, and $\langle a,b\rangle_{G_k}:=a^\top G_k b$. For stable discrete injection (\cref{sec:method_discrete}), we normalize the direction and let the scalar schedule $\eta_k$ exclusively control the step size:
\begin{equation}
u_{\text{hat}}=\frac{u_k^\star}{\|u_k^\star\|+\epsilon},
\qquad
x_{k-1}=x_{k-1}^{\text{clean}}+\eta_k\,u_{\text{hat}}.
\label{eq:proj_norm_update}
\end{equation}
Consistent with prior practice and our discrete analysis (\cref{eq:w2_bound}), we use a \emph{late-window} schedule (e.g., last $20\%$ of steps)~\cite{dai_advdiff_2024} to reduce early trajectory drift.

\subsection{Metric Choices and Stability}
\label{sec:method_metric}
Our practical implementation (\cref{alg:dpac}) requires computing an inner product between an $x$-space gradient ($w_k\propto\nabla_x\ell$) and a score $s_k$.
This is an approximation of the theoretically pure tangential projection, so we consider two computationally cheap metric choices $G_t$: (A) Identity (Euclidean): $G_t=I$.
This computes a standard Euclidean inner product, treating the two fields as commensurate. (B) Noise-Scaled: $G_t=(1-\alpha_t)^{-1}I$.
Here, $\alpha_t$ denotes the cumulative signal power at time $t$ (i.e., $\bar{\alpha}_t$ in DDPM~\cite{ho_ddpm_2020} notation).
This choice is motivated by the scaling of the score in diffusion theory, weighting by the inverse noise variance $(1-\alpha_t)^{-1}$.
We empirically evaluate its effect in \cref{sec:ablation}.

The raw gradient $w_k=\mathcal{G}_{k-1}(x_{k-1}^{\text{clean}})$ can have an arbitrarily large magnitude.
Directly injecting this vector (even after projection) can lead to numerical instability and sample collapse (see \cref{sec:ablation}).
A simple clipping approach is insufficient because the perturbation magnitude remains coupled to the raw gradient scale.
DPAC therefore uses a critical Project-then-Normalize sequence:
we first compute the unnormalized projected direction $u_k^\star$ via \cref{eq:score_proj_method}, then normalize to a unit vector $u_{\text{hat}}$ via \cref{eq:proj_norm_update}.
This disentangles direction (set by projection) from magnitude (set solely by $\eta_k$), preventing collapse in naive guidance.

\subsection{Stable Injection via Denoise-then-Perturb}
\label{sec:method_discrete}
While the core theory is written in terms of modifying the reverse-time drift, directly injecting a drift term in discrete samplers is numerically unstable in practice:
as $g_t\Delta t\to 0$, drift-level modifications yield vanishing control signals.
Instead, we implement DPAC using a robust Denoise-then-Perturb (PGD-style) mechanism. For each step $k$, we first take a standard denoising step with the base sampler $\Phi_{\text{base}}$, and then inject a perturbation in $x$-space:
\begin{align}
\label{eq:pgd_denoise}
x_{k-1}^{\text{clean}} &= \Phi_{\text{base}}(x_k, s_k, t_k), \\
\label{eq:pgd_perturb}
x_{k-1} &= x_{k-1}^{\text{clean}} + \eta_k \cdot u_{\text{hat}},
\end{align}
where $u_{\text{hat}}$ is the normalized projected direction from \cref{eq:proj_norm_update}.
This implementation preserves the \emph{direction} implied by our theory (projection removes the score-parallel, density-changing component), while re-parameterizing the \emph{magnitude} of the discrete update through the schedule $\eta_k$.
As a result, DPAC realizes an effective $O(1)$ displacement per active step, independently of the small factor $g_t\Delta t$ that would scale a drift-level control.

\begin{algorithm}[t]
 \caption{DPAC-Guided Sampling (Denoise-then-Perturb)}
 \label{alg:dpac}
 \begin{algorithmic}[1]
    \Require initial noise $x_K$, timesteps $\{t_k\}_{k=K}^1$, ground-truth condition $y_{\text{gt}}$
    \Require score model $s_\theta(\cdot,t\mid y_{\text{gt}})$, base sampler $\Phi_{\text{base}}$
    \Require gradient oracle $\mathcal{G}_{k-1}$, schedule $\{\eta_k\}_{k=K}^1$, metric choice $G_k$
    \Ensure $x_0$
    \State $x_k \gets x_K$
    \For{$k = K$ \textbf{down to} $1$}
        \State $s_k \gets s_\theta(x_k, t_k, y_{\text{gt}})$
        \State $x_{k-1}^{\text{clean}} \gets \Phi_{\text{base}}(x_k, s_k, t_k)$
        \If{$\eta_k > 0$}
            \State $w_k \gets \mathcal{G}_{k-1}(x_{k-1}^{\text{clean}})$
            \State $G_k \gets I \ \textbf{or}\ (1-\alpha_k)^{-1}I$
            \State $u_k^\star \gets w_k - \frac{\langle w_k, s_k \rangle_{G_k}}{\langle s_k, s_k \rangle_{G_k} + \epsilon}\, s_k$
            \State $u_{\text{hat}} \gets u_k^\star / (\|u_k^\star\| + \epsilon)$
            \State $x_{k-1} \gets x_{k-1}^{\text{clean}} + \eta_k \cdot u_{\text{hat}}$
        \Else
            \State $x_{k-1} \gets x_{k-1}^{\text{clean}}$
        \EndIf
        \State $x_k \gets x_{k-1}$
    \EndFor
    \State \Return $x_0$
 \end{algorithmic}
\end{algorithm}

\subsection{Complexity and Implementation Notes}
\label{sec:method_complexity}
Let $C_{\text{score}}$ be the cost of one score evaluation and $C_{\text{grad}}$ be the cost of one gradient (sensitivity) query.

Per active step, DPAC adds:
(i) one sensitivity query $w_k=\mathcal{G}_{k-1}(x_{k-1}^{\text{clean}})$ (\cref{eq:pgd_gradient});
(ii) one metric inner product and projection (\cref{eq:score_proj_method});
(iii) a normalization and a PGD perturbation (\cref{eq:proj_norm_update}).
Since the sensitivity query is the dominant cost (forward/backward through the classifier, potentially including a VAE decode),
the per-step overhead during the active window is roughly $C_{\text{grad}} + C_{\text{score}}$.

Because the final implementation (\cref{alg:dpac}) injects a normalized direction rather than a drift-level control, the Girsanov identity from \cref{sec:theory} no longer applies directly.
We therefore report \emph{Cumulative Perturbation Energy} (CPE) as a practical proxy that reflects the magnitude of the underlying (pre-normalization) guidance field:
\begin{equation}
\widehat{\mathcal{E}}_{\text{CPE}} = \tfrac{1}{2} \sum_{k} \|\eta_k u_k\|_2^2,
\label{eq:cpe}
\end{equation}
where $u_k$ is the unnormalized intervention vector:
$u_k=u_k^\star$ for DPAC (\cref{eq:score_proj_method}) and $u_k=w_k$ for AdvDiff (\cref{eq:pgd_gradient}).
We share the noise $\{\epsilon_k\}$ across methods (common random numbers) to reduce variance in $\widehat{\mathcal{E}}_{\text{CPE}}$ and to align comparisons with the coupling used in our discrete analysis.

The Project-then-Normalize scheme and the PGD-style update prevent catastrophic numerical explosion caused by raw gradients.
Under score/metric/sensitivity estimation errors, the direction remains bounded and the step size remains controlled by $\eta_k$. Unless stated otherwise, we use PGD injection, a linear $\eta_k$ ramp over the last $20\%$ of steps, and $G_t=I$ for the main DPAC result.
\section{Experiments}
\label{sec:experiments}

\begin{figure*}[t!]
    \centering
    \begin{subfigure}[t]{0.32\linewidth}
        \centering
        \includegraphics[width=\linewidth]{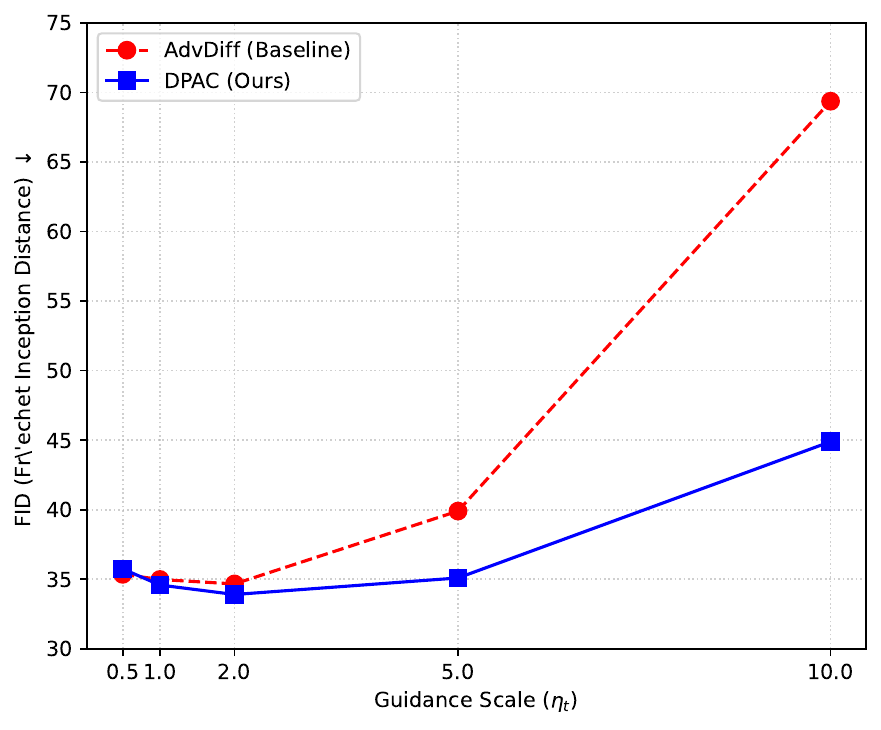}
        \caption{Stability (FID vs. Scale)}
        \label{fig:tradeoff_fid_scale}
    \end{subfigure}\hfill
    \begin{subfigure}[t]{0.32\linewidth}
        \centering
        \includegraphics[width=\linewidth]{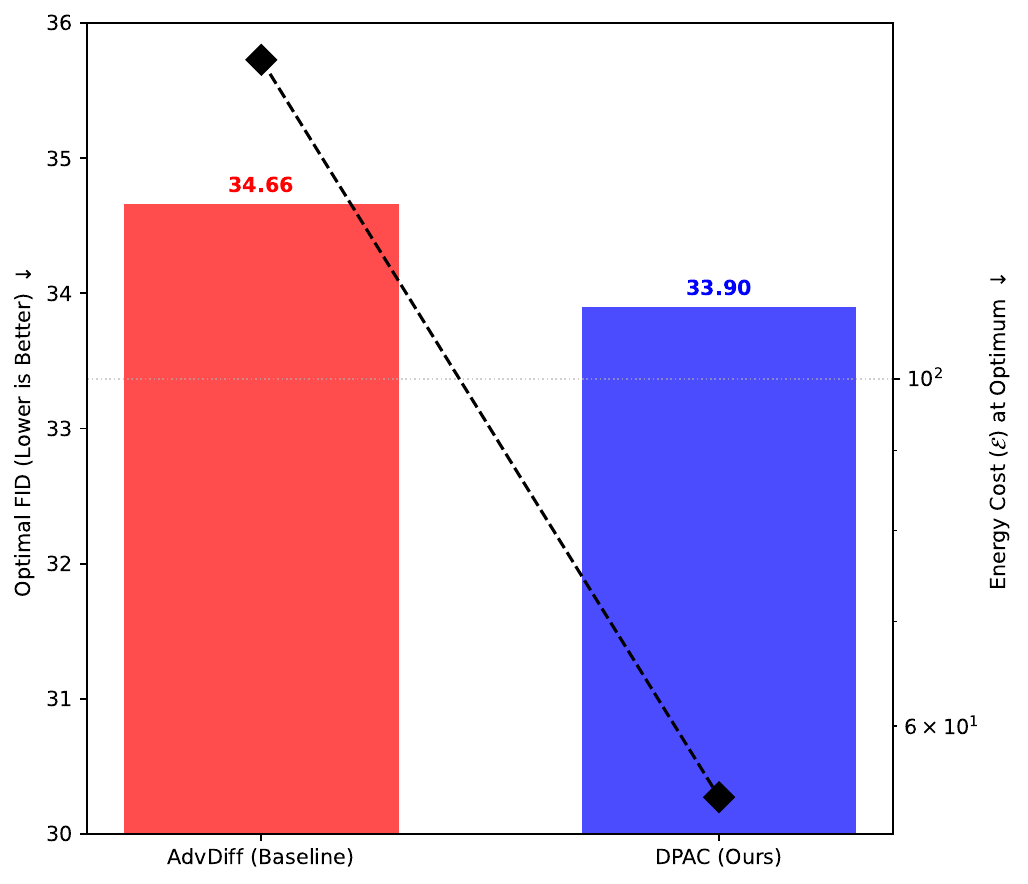}
        \caption{Peak Effectiveness \& Efficiency}
        \label{fig:tradeoff_optimal}
    \end{subfigure}\hfill
    \begin{subfigure}[t]{0.32\linewidth}
        \centering
        \includegraphics[width=\linewidth]{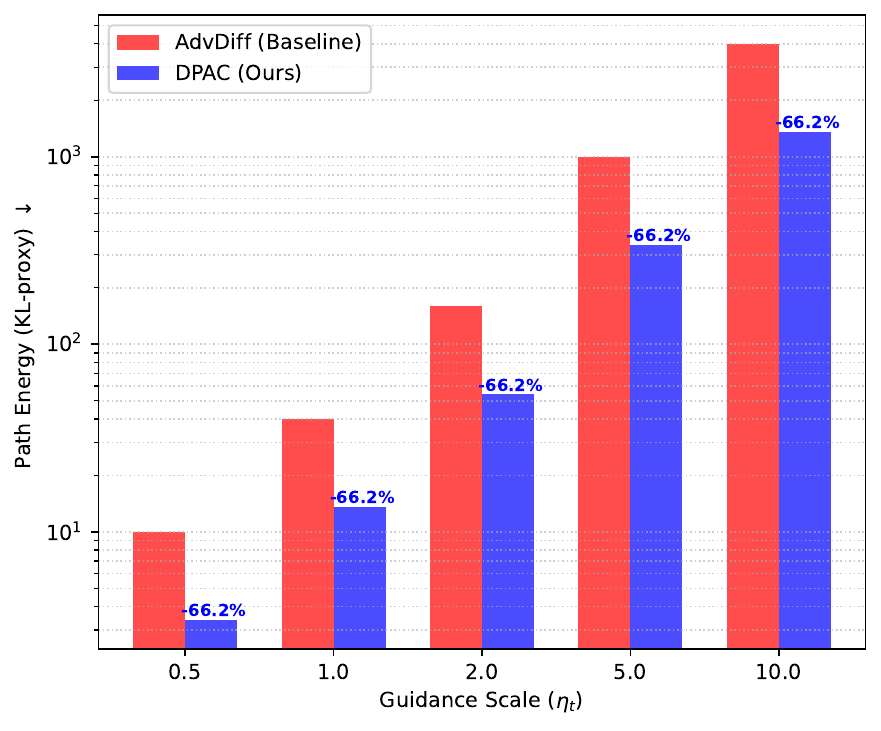}
        \caption{Theoretical Validation (Energy)}
        \label{fig:tradeoff_energy_bar}
    \end{subfigure}
    \caption{
    \textbf{Quantitative validation of DPAC on ImageNet-100 (200 steps).}
    \textbf{(a) Stability:} AdvDiff (red) suffers catastrophic FID collapse (39.9 $\to$ 69.37) at high guidance scales ($\eta_k$). DPAC (blue) remains robustly stable.
    \textbf{(b) Effectiveness \& Efficiency:} A direct comparison of the best FID each method achieved. DPAC (blue) achieves a superior peak fidelity (FID 33.90) while using only one-third of the energy ($\mathcal{E}$=54.0) that AdvDiff (red) required for its worse optimum (FID 34.66 at $\mathcal{E}$=160.0).
    \textbf{(c) Theoretical Validation:} At all scales, DPAC consistently uses $\approx$66\% less energy, empirically validating our theory.
    }
    \label{fig:main_quantitative_results}
\end{figure*}

Our experiments are designed to empirically validate our core theoretical claims:
(1) that standard gradient guidance, as exemplified by methods like AdvDiff~\cite{dai_advdiff_2024}, injects a score-parallel component that destabilizes the sampling trajectory and leads to distributional collapse (high FID);
and (2) that our proposed DPAC (tangential control) mitigates this issue by projecting away this harmful component.
We emphasize that DPAC is a minimal, projection-based instantiation of our analysis to test the predicted mechanism rather than to engineer a new sampler, and our evaluation focuses on controlled, like-for-like comparisons that isolate the control direction and its distributional impact.

\subsection{Experimental Setup}
\label{sec:setup}
We conduct a direct, controlled comparison between two guidance strategies, both implemented using an identical Denoise-then-Perturb (PGD-style) injection mechanism.
This isolates the impact of the \emph{control direction} itself.
\begin{itemize}
   \item AdvDiff (Baseline): We re-implement the core guidance mechanism of AdvDiff~\cite{dai_advdiff_2024}.
    This method uses the raw classifier gradient $w_k = \nabla_{x_{k-1}^{\text{clean}}} \ell_{\text{tar}}  (x_{k-1}^{\text{clean}})$ as the control direction (i.e., gradient ascent to maximize loss).
    This injects a score-parallel component (normal to iso-density surfaces), which our theory predicts will distort the distribution.
    \item DPAC (Ours): The tangential control method from \cref{sec:method}.
    It uses the $G_k$-inner-product projection of the same ascent direction $w_k$ onto the subspace orthogonal to the score $s_k$, i.e., $u_{\text{hat}}\propto \Pi_\perp^{(s_k,G_k)}(w_k)$ as in \cref{eq:score_proj}.
    We test two variants for $G_k$: the identity ($G_k\!=\!I$) and the noise-scaled metric ($G_k\!=\!(1-\alpha_k)^{-1}I$).
\end{itemize}
Among existing approaches, AdvDiff serves as the closest baseline that performs in-trajectory classifier-gradient guidance for unrestricted adversarial example generation; in contrast, DiffPGD~\cite{xue_diffusion-based_2023} and AdvDiffuser~\cite{chen_advdiffuser_2023} apply diffusion priors outside the reverse dynamics and are thus not directly comparable in our controlled setting.

We use the pre-trained Latent Diffusion Model (LDM)~\cite{robin_high_2022} checkpoint (cin256-v2) on ImageNet-1K.
For efficient evaluation, all experiments are conducted on the ImageNet-100 (IN-100) subset (classes 0-99).
We use a pre-trained ResNet50~\cite{he_deep_2016} (V1 legacy, `pretrained=True') as the classifier $v_\phi$ for both guidance and ASR evaluation.
The CFG scale for $y_{\text{gt}}$ is fixed to 3.0.
Unless otherwise stated, the target is $y_{\text{tar}}=(y_{\text{gt}}+1)\!\!\pmod{1000}$;
ASR is computed on the full 1K head (R50-V1), while FID/IS use IN-100 validation statistics.
All sampling uses DDIM with 200 steps (main results) or 50 steps (ablation) on 1,300 samples (13 per class).
While 1,300 samples are limited for absolute FID/IS estimation, our primary goal is the \emph{relative} comparison between guidance strategies. To this end, we use shared-noise coupling across methods (as discussed in \cref{sec:method_complexity}).
This isolates the impact of the control $u_k$ from stochastic sampling variance, aligning with our theoretical bounds (\cref{eq:w2_bound}) and ensuring a robust relative comparison.
%
The control signal is injected with a linear ramp-up of $\eta_k$.

We evaluate three key aspects:
(i) Effectiveness (ASR, R50-V1): Attack Success Rate—the percentage of samples classified as $y_{\text{tar}}$ by the ResNet-50 (V1) classifier.
(ii) Fidelity \& Stability (FID/IS): We use Fr\'echet Inception Distance (FID)~\cite{heusel_GANs_2017} (lower is better) and Inception Score (IS)~\cite{salimans_improved_2016} (higher is better).
(iii) Efficiency (CPE): We report CPE, as defined in \cref{eq:cpe} (\cref{sec:method_complexity}). This metric measures the energy of the unnormalized control vector $u_k$ and aligns with our theoretical analysis (cf. \cref{sec:core_validation}).

\begin{table*}[t!]
    \centering
    \caption{\textbf{Quantitative comparison on ImageNet-100 (200 steps) across all guidance scales.}
    DPAC achieves a superior optimal FID (33.90) at one-third the energy cost of AdvDiff's optimum (FID 34.66).
    Crucially, AdvDiff suffers catastrophic FID collapse (39.9 $\to$ 69.37) at high scales, while DPAC remains stable (44.89).
    At every scale, DPAC is more efficient, consuming $\approx$66\% less CPE.
    }
    \label{tab:main_results}
    \resizebox{\linewidth}{!}{%
    \begin{tabular}{l|c|cccc|cccc}
        \toprule
        & & \multicolumn{4}{c|}{\textbf{AdvDiff (Baseline)}} & \multicolumn{4}{c}{\textbf{DPAC (Ours, $G_k=I$)}} \\
        \textbf{Metric} & Scale ($\eta_k$) & ASR (R50\textendash V1) ($\uparrow$) & FID ($\downarrow$) & IS ($\uparrow$) & CPE ($\downarrow$) & ASR (R50\textendash V1) ($\uparrow$) & FID ($\downarrow$) & IS ($\uparrow$) & CPE ($\downarrow$) \\
        \midrule
        & 0.5 & 97.1\% & 35.35 & 33.30 & 10.0 & 50.1\% & 35.74 & \textbf{34.29} & \textbf{3.38} \\
        & 1.0 & \textbf{100.0\%} & 34.98 & 32.19 & 40.0 & 97.9\% & 34.58 & \textbf{33.04} & \textbf{13.50} \\
        Optimal FID $\to$ & 2.0 & \textbf{100.0\%} & 34.66 & 30.72 & 160.0 & \textbf{100.0\%} & \textbf{33.90} & \textbf{31.48} & \textbf{54.02} \\
        \& Stability & 5.0 & 100.0\% & 39.90 & 24.77 & 1000.0 & 99.9\% & 35.09 & \textbf{28.46} & \textbf{337.61} \\
        & 10.0 & 100.0\% & \textbf{69.37} & 20.62 & 4000.0 & 99.6\% & \textbf{44.89} & \textbf{22.86} & \textbf{1350.43} \\
        \bottomrule
    \end{tabular}%
    }
\end{table*}

\begin{figure*}[t!]
    \centering
    \includegraphics[width=\linewidth]{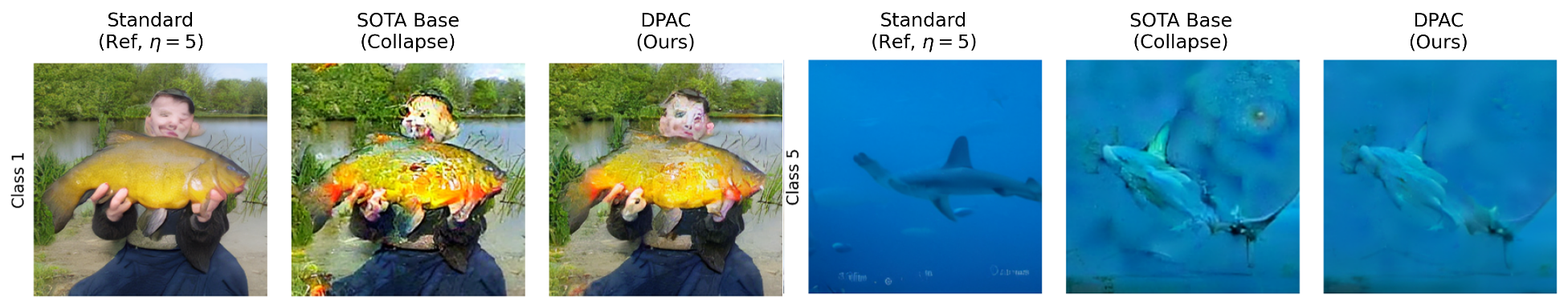}
    \caption{
        \textbf{Qualitative evidence of collapse and its mitigation (ImageNet-100, S200).}
        Columns show \emph{(left)} DPAC at a moderate scale ($\eta{=}5$) as a non-collapsed reference,
        \emph{(middle)} AdvDiff at high scale ($\eta{=}10$; Tab.~\ref{tab:main_results}, $\eta{=}10$ row) exhibiting severe collapse,
        and \emph{(right)} DPAC at the same high scale ($\eta{=}10$) preserving coherent structure.
    }
    \label{fig:qual_collapse}
\end{figure*}

\subsection{Core Validation}
\label{sec:core_validation}

Our S200 results (Tab.~\ref{tab:main_results} and Fig.~\ref{fig:main_quantitative_results}) validate our claims.
Fig.~\ref{fig:qual_collapse} provides a qualitative counterpart: it contrasts AdvDiff at $\eta{=}10$ against DPAC at $\eta{=}10$, using DPAC at $\eta{=}5$ as a non-collapsed visual reference (shared noise across columns).

At high guidance ($\eta{=}10$), AdvDiff collapses (FID $39.9\!\to\!69.37$), consistent with distributional drift induced by the score-parallel component, whereas DPAC remains stable (FID $44.89$).
This failure mode is also visible qualitatively (Fig.~\ref{fig:qual_collapse}): the baseline exhibits severe color/texture corruption and structural distortion, while DPAC avoids catastrophic artifacts and preserves coherent structure.

DPAC reduces energy by $\approx66\%$ across scales (Fig.~\ref{fig:tradeoff_energy_bar}), matching the geometric prediction that removing the score-parallel share reduces required control magnitude.
Moreover, DPAC attains a better peak FID ($33.90$) at substantially lower energy (54 vs. 160), showing that stability and peak fidelity can improve simultaneously when the control direction is corrected.

\subsection{Ablation Studies}
\label{sec:ablation}
We ablate the choice of the inner product matrix $G_k$ used for the orthogonal projection, comparing our default $G_k=I$ (identity) against the theoretically-motivated $G_k=(1-\alpha_k)^{-1}I$ (noise\_scaled).
As shown in \cref{tab:ablation_metric}, the choice of metric has a negligible impact on the final results, yielding a nearly identical ASR--FID trade-off.
This suggests that while $G_k=(1-\alpha_k)^{-1}I$ is geometrically precise, the practical choice $G_k=I$ is sufficient and simpler.

\noindent \textbf{Further Analysis.}
The Supplementary Material provides additional experiments and analyses, including (i) statistical significance with 95\% confidence intervals and a negative-control study, (ii) stability at high guidance across varying DDIM steps $N\in\{50,100,150,200\}$, (iii) generalization across guidance classifier backbones, (iv) an ablation of guidance-window strategies (Early/Full/Late), and (v) extensive qualitative comparisons in both extreme and standard regimes.
These results further support the proposed mechanism and robustness beyond the main setting.

\begin{table}[t]
    \centering
    \caption{\textbf{Ablation on Inner Product Metric $G_k$ (S50).}
    Both metrics yield a nearly identical ASR–FID trade-off, validating the simpler choice $G_k=I$.
    }
    \label{tab:ablation_metric}
    \begin{tabular}{l|c|cc}
        \toprule
        \textbf{Method} & Scale ($\eta_k$) & ASR (R50\textendash V1) ($\uparrow$) & FID ($\downarrow$) \\
        \midrule
        DPAC ($G_k=I$) & 5.0 & 93.3\% & 34.46 \\
        DPAC ($G_k=(1 - \alpha_k)^{-1}I$) & 5.0 & 92.6\% & 34.54 \\
        \midrule
        DPAC ($G_k=I$) & 10.0 & 99.3\% & 35.05 \\
        DPAC ($G_k=(1 - \alpha_k)^{-1}I$) & 10.0 & 99.3\% & 34.74 \\
        \bottomrule
    \end{tabular}%
\end{table}
\section{Related Works}
\label{sec:related_works}

Classical $\ell_p$-bounded attacks such as FGSM~\cite{ian_explaining_2015} and PGD~\cite{madry_towards_2018} optimize a perturbation $\delta$ for a given input $x$.
Beyond norm-bounded threats, unrestricted or generative adversarial examples synthesize adversarial images from scratch using generative models (e.g., AC-GAN)~\cite{song_constructing_2018} or produce perturbations via GANs~\cite{xiao_generating_2018}.

Diffusion guidance methods like Classifier Guidance (CG)~\cite{dhariwal_diffusion_2021} and Classifier-Free Guidance (CFG)~\cite{ho_classifier_2022} steer sampling using classifier gradients or unconditional scores, respectively, to trade off fidelity and diversity.
While both mechanisms are effective, increasing the guidance scale is known to improve alignment at the expense of sample quality (e.g., FID), revealing an intrinsic guidance--fidelity tension~\cite{dhariwal_diffusion_2021,ho_classifier_2022}.

A body of work integrates PGD-style optimization with diffusion to enhance realism and controllability.
For example, Diff-PGD~\cite{xue_diffusion-based_2023} leverages a diffusion prior to keep attacks on-manifold, while AdvDiffuser~\cite{chen_advdiffuser_2023} conducts step-wise optimization to perturb less-salient regions.

AdvDiff~\cite{dai_advdiff_2024} stands as the primary baseline, generating UAEs by injecting raw classifier gradients into the continuous diffusion sampling process.
While this method suffers from a severe fidelity collapse that prior work treats as an empirical trade-off, our work (DPAC) diagnoses this instability and mitigates it via tangential projection.
Other methods like Diff-PGD~\cite{xue_diffusion-based_2023} and AdvDiffuser~\cite{chen_advdiffuser_2023} are not directly comparable, as they are image-editing AE methods requiring source images, unlike our from-scratch UAE generation setting.

Path Integral Sampler~\cite{zhang2021path} formulates sampling from complex targets through an SOC lens and derives a $\nabla\log(\cdot)$-type controller via a path-integral representation.
Stochastic Control Guidance~\cite{huang_scg_rule_guided_diffusion} applies an SOC-inspired, plug-and-play guidance mechanism to incorporate non-differentiable constraints into diffusion sampling. These SOC/neural-sampler perspectives are complementary to ours: DPAC focuses specifically on \emph{adversarial} guidance and shows that, for a prescribed first-order classification gain, removing the score-parallel (density-changing) component yields a distribution-preserving direction that reduces the required control magnitude.
\section{Discussion}
In practice, the exact $p_t$-tangential projector is intractable; DPAC removes only the score-parallel component and thus acts as a first-order surrogate, so residual density change can remain with approximate scores and finite-step discretization.
This motivates the late-window schedule and project-then-normalize Denoise-then-Perturb update in \cref{sec:method}; since this injection breaks the exact path--KL identity, we report CPE as a proxy and use shared-noise coupling to reduce variance in \cref{sec:experiments}.
We also observe that gains can shrink for stronger classifiers (see Supplementary Material), which is consistent with their adversarial directions being closer to tangential; beyond UAE generation, the same projection principle can be applied to other conditional samplers (e.g., CFG-style guidance) and to velocity/flow-based generative models by projecting the corresponding guidance field.
Finally, because DPAC enables high-fidelity targeted UAEs, deployments should include provenance signals and audit logging to mitigate misuse.

\section{Conclusion}
We connect adversarial diffusion guidance to distributional distortion measured by path--KL (energy), show that under a fixed first-order gain the $p_t$-tangential component yields the minimum-energy direction, and that its discrete counterpart removes the leading score-parallel term in a Wasserstein bound.
Instantiated with a score-orthogonal projector and a stable denoise-then-perturb update, DPAC prevents catastrophic collapse in our setting and improves the FID--CPE trade-off on ImageNet-100, supporting energy minimization as a practical principle for robust guidance.

\bibliographystyle{splncs04}
\bibliography{main}

@String(CVPR  = {IEEE Conf. Comput. Vis. Pattern Recog.})

@String(ICCV  = {Int. Conf. Comput. Vis.})

@String(ECCV  = {Eur. Conf. Comput. Vis.})

@String(NeurIPS = {Adv. Neural Inform. Process. Syst.})

@String(ICML  = {Int. Conf. Mach. Learn.})

@String(ICLR  = {Int. Conf. Learn. Represent.})

@String(AAAI  = {AAAI})

@String(IJCAI = {IJCAI})

@String(CVPR  = {CVPR})

@String(ICCV  = {ICCV})

@String(ECCV  = {ECCV})

@String(NeurIPS = {NeurIPS})

@String(ICML  = {ICML})

@String(ICLR  = {ICLR})

@inproceedings{zhang2021path,
  author    = {Qinsheng Zhang and Yongxin Chen},
  title     = {Path Integral Sampler: a stochastic control approach for sampling},
  booktitle = ICLR,
  year      = {2022}
}

@inproceedings{heusel_GANs_2017,
author = {Heusel, Martin and Ramsauer, Hubert and Unterthiner, Thomas and Nessler, Bernhard and Hochreiter, Sepp},
title = {GANs trained by a two time-scale update rule converge to a local nash equilibrium},
year = {2017},
isbn = {9781510860964},
publisher = {Curran Associates Inc.},
address = {Red Hook, NY, USA},
booktitle = {Proceedings of the 31st International Conference on Neural Information Processing Systems},
pages = {6629–6640},
numpages = {12},
location = {Long Beach, California, USA},
series = {NIPS'17}
}

@inproceedings{ho_ddpm_2020,
  author       = {Jonathan Ho and
                  Ajay Jain and
                  Pieter Abbeel},
  editor       = {Hugo Larochelle and
                  Marc'Aurelio Ranzato and
                  Raia Hadsell and
                  Maria{-}Florina Balcan and
                  Hsuan{-}Tien Lin},
  title        = {Denoising Diffusion Probabilistic Models},
  booktitle    = {Advances in Neural Information Processing Systems 33: Annual Conference
                  on Neural Information Processing Systems 2020, NeurIPS 2020, December
                  6-12, 2020, virtual},
  year         = {2020},
  url          = {https://proceedings.neurips.cc/paper/2020/hash/4c5bcfec8584af0d967f1ab10179ca4b-Abstract.html},
  bibsource    = {dblp computer science bibliography, https://dblp.org}
}

@inproceedings{song_ddim_2021,
  author       = {Jiaming Song and
                  Chenlin Meng and
                  Stefano Ermon},
  title        = {Denoising Diffusion Implicit Models},
  booktitle    = {9th International Conference on Learning Representations, {ICLR} 2021,
                  Virtual Event, Austria, May 3-7, 2021},
  publisher    = {OpenReview.net},
  year         = {2021},
  url          = {https://openreview.net/forum?id=St1giarCHLP},
  bibsource    = {dblp computer science bibliography, https://dblp.org}
}

@article{ho_classifier_2022,
  title={Classifier-Free Diffusion Guidance},
  author={Jonathan Ho},
  journal={ArXiv},
  year={2022},
  volume={abs/2207.12598},
  url={https://api.semanticscholar.org/CorpusID:249145348}
}

@inproceedings{huang_scg_rule_guided_diffusion,
author = {Huang, Yujia and Ghatare, Adishree and Liu, Yuanzhe and Hu, Ziniu and Zhang, Qinsheng and Sastry, Chandramouli S and Gururani, Siddharth and Oore, Sageev and Yue, Yisong},
title = {Symbolic music generation with non-differentiable rule guided diffusion},
year = {2024},
publisher = {JMLR.org},
booktitle = {Proceedings of the 41st International Conference on Machine Learning},
articleno = {796},
numpages = {26},
location = {Vienna, Austria},
series = {ICML'24}
}

@article{aenderson_reverse_1982,
title = {Reverse-time diffusion equation models},
journal = {Stochastic Processes and their Applications},
volume = {12},
number = {3},
pages = {313-326},
year = {1982},
issn = {0304-4149},
doi = {https://doi.org/10.1016/0304-4149(82)90051-5},
author = {Brian D.O. Anderson}
}

@inproceedings{song_score_2021,
  author       = {Yang Song and
                  Jascha Sohl{-}Dickstein and
                  Diederik P. Kingma and
                  Abhishek Kumar and
                  Stefano Ermon and
                  Ben Poole},
  title        = {Score-Based Generative Modeling through Stochastic Differential Equations},
  booktitle    = {9th International Conference on Learning Representations, {ICLR} 2021,
                  Virtual Event, Austria, May 3-7, 2021},
  publisher    = {OpenReview.net},
  year         = {2021},
  url          = {https://openreview.net/forum?id=PxTIG12RRHS},
  bibsource    = {dblp computer science bibliography, https://dblp.org}
}

@article{robin_high_2022,
  title={High-Resolution Image Synthesis with Latent Diffusion Models},
  author={Robin Rombach and A. Blattmann and Dominik Lorenz and Patrick Esser and Bj{\"o}rn Ommer},
  journal={2022 IEEE/CVF Conference on Computer Vision and Pattern Recognition (CVPR)},
  year={2021},
  pages={10674-10685},
  url={https://api.semanticscholar.org/CorpusID:245335280}
}

@inproceedings{dhariwal_diffusion_2021,
author = {Dhariwal, Prafulla and Nichol, Alex},
title = {Diffusion models beat GANs on image synthesis},
year = {2021},
isbn = {9781713845393},
publisher = {Curran Associates Inc.},
address = {Red Hook, NY, USA},
booktitle = {Proceedings of the 35th International Conference on Neural Information Processing Systems},
articleno = {672},
numpages = {15},
series = {NIPS '21}
}

@inproceedings{dai_advdiff_2024,
author = {Dai, Xuelong and Liang, Kaisheng and Xiao, Bin},
title = {AdvDiff: Generating Unrestricted Adversarial Examples Using Diffusion Models},
year = {2024},
isbn = {978-3-031-72951-5},
publisher = {Springer-Verlag},
address = {Berlin, Heidelberg},
url = {https://doi.org/10.1007/978-3-031-72952-2_6},
doi = {10.1007/978-3-031-72952-2_6},
booktitle = {Computer Vision – ECCV 2024: 18th European Conference, Milan, Italy, September 29–October 4, 2024, Proceedings, Part XLVI},
pages = {93–109},
numpages = {17},
location = {Milan, Italy}
}

@INPROCEEDINGS{chen_advdiffuser_2023,
  author={Chen, Xinquan and Gao, Xitong and Zhao, Juanjuan and Ye, Kejiang and Xu, Cheng-Zhong},
  booktitle={2023 IEEE/CVF International Conference on Computer Vision (ICCV)}, 
  title={AdvDiffuser: Natural Adversarial Example Synthesis with Diffusion Models}, 
  year={2023},
  volume={},
  number={},
  pages={4539-4549},
  doi={10.1109/ICCV51070.2023.00421}}

@inproceedings{xue_diffusion-based_2023,
author = {Xue, Haotian and Araujo, Alexandre and Hu, Bin and Chen, Yongxin},
title = {Diffusion-based adversarial sample generation for improved stealthiness and controllability},
year = {2023},
publisher = {Curran Associates Inc.},
address = {Red Hook, NY, USA},
booktitle = {Proceedings of the 37th International Conference on Neural Information Processing Systems},
articleno = {129},
numpages = {28},
location = {New Orleans, LA, USA},
series = {NIPS '23}
}

@Inbook{kallianpur_girsanov_2000,
author="Kallianpur, Gopinath
and Karandikar, Rajeeva L.",
title="Girsanov's Theorem",
bookTitle="Introduction to Option Pricing Theory",
year="2000",
publisher="Birkh{\"a}user Boston",
address="Boston, MA",
pages="95--101",
isbn="978-1-4612-0511-1",
doi="10.1007/978-1-4612-0511-1_5",
url="https://doi.org/10.1007/978-1-4612-0511-1_5"
}

@book{Oksendal_SDE_1992,
author = {Oksendal, Bernt},
title = {Stochastic differential equations (3rd ed.): an introduction with applications},
year = {1992},
isbn = {3387533354},
publisher = {Springer-Verlag},
address = {Berlin, Heidelberg}
}

@Inbook{molly_talagrand_2002,
author="Molloy, Michael
and Reed, Bruce",
title="A Closer Look at Talagrand's Inequality",
bookTitle="Graph Colouring and the Probabilistic Method",
year="2002",
publisher="Springer Berlin Heidelberg",
address="Berlin, Heidelberg",
pages="231--236",
isbn="978-3-642-04016-0",
doi="10.1007/978-3-642-04016-0_20",
url="https://doi.org/10.1007/978-3-642-04016-0_20"
}

@ARTICLE{bhatia_helmholtz_2013,
  author={Bhatia, Harsh and Norgard, Gregory and Pascucci, Valerio and Bremer, Peer-Timo},
  journal={IEEE Transactions on Visualization and Computer Graphics}, 
  title={The Helmholtz-Hodge Decomposition—A Survey}, 
  year={2013},
  volume={19},
  number={8},
  pages={1386-1404},
  doi={10.1109/TVCG.2012.316}}

@inproceedings{chen_neural_2018,
author = {Chen, Ricky T. Q. and Rubanova, Yulia and Bettencourt, Jesse and Duvenaud, David},
title = {Neural ordinary differential equations},
year = {2018},
publisher = {Curran Associates Inc.},
address = {Red Hook, NY, USA},
booktitle = {Proceedings of the 32nd International Conference on Neural Information Processing Systems},
pages = {6572–6583},
numpages = {12},
location = {Montr\'{e}al, Canada},
series = {NIPS'18}
}

@INPROCEEDINGS{he_deep_2016,
  author={He, Kaiming and Zhang, Xiangyu and Ren, Shaoqing and Sun, Jian},
  booktitle={2016 IEEE Conference on Computer Vision and Pattern Recognition (CVPR)}, 
  title={Deep Residual Learning for Image Recognition}, 
  year={2016},
  volume={},
  number={},
  pages={770-778},
  doi={10.1109/CVPR.2016.90}}

@inproceedings{salimans_improved_2016,
author = {Salimans, Tim and Goodfellow, Ian and Zaremba, Wojciech and Cheung, Vicki and Radford, Alec and Chen, Xi},
title = {Improved techniques for training GANs},
year = {2016},
isbn = {9781510838819},
publisher = {Curran Associates Inc.},
address = {Red Hook, NY, USA},
booktitle = {Proceedings of the 30th International Conference on Neural Information Processing Systems},
pages = {2234–2242},
numpages = {9},
location = {Barcelona, Spain},
series = {NIPS'16}
}

@inproceedings{ian_explaining_2015,
  author       = {Ian J. Goodfellow and
                  Jonathon Shlens and
                  Christian Szegedy},
  editor       = {Yoshua Bengio and
                  Yann LeCun},
  title        = {Explaining and Harnessing Adversarial Examples},
  booktitle    = {3rd International Conference on Learning Representations, {ICLR} 2015,
                  San Diego, CA, USA, May 7-9, 2015, Conference Track Proceedings},
  year         = {2015},
  url          = {http://arxiv.org/abs/1412.6572},
  bibsource    = {dblp computer science bibliography, https://dblp.org}
}

@inproceedings{madry_towards_2018,
  author       = {Aleksander Madry and
                  Aleksandar Makelov and
                  Ludwig Schmidt and
                  Dimitris Tsipras and
                  Adrian Vladu},
  title        = {Towards Deep Learning Models Resistant to Adversarial Attacks},
  booktitle    = {6th International Conference on Learning Representations, {ICLR} 2018,
                  Vancouver, BC, Canada, April 30 - May 3, 2018, Conference Track Proceedings},
  publisher    = {OpenReview.net},
  year         = {2018},
  url          = {https://openreview.net/forum?id=rJzIBfZAb},
  bibsource    = {dblp computer science bibliography, https://dblp.org}
}

@inproceedings{xiao_generating_2018,
author = {Xiao, Chaowei and Li, Bo and Zhu, Jun-Yan and He, Warren and Liu, Mingyan and Song, Dawn},
title = {Generating adversarial examples with adversarial networks},
year = {2018},
isbn = {9780999241127},
publisher = {AAAI Press},
booktitle = {Proceedings of the 27th International Joint Conference on Artificial Intelligence},
pages = {3905–3911},
numpages = {7},
location = {Stockholm, Sweden},
series = {IJCAI'18}
}

@inproceedings{song_constructing_2018,
  title={Constructing Unrestricted Adversarial Examples with Generative Models},
  author={Yang Song and Rui Shu and Nate Kushman and Stefano Ermon},
  booktitle={Neural Information Processing Systems},
  year={2018},
  url={https://api.semanticscholar.org/CorpusID:52309169}
}
\end{document}